\documentclass[letterpaper, 10 pt, conference]{ieeeconf}  
\IEEEoverridecommandlockouts                              

\usepackage{graphicx}
\usepackage{hyperref}
\usepackage{xcolor}
\usepackage{color}
\usepackage{balance}

\title{\LARGE \bf
Palpation Alters Auditory Pain Expressions with Gender-Specific Variations in Robopatients
}

\author{Chapa Sirithunge$^{1}$, Yue Xie$^{1}$, Saitarun Nadipineni$^{2}$, Fumiya Iida$^{1}$ and Thilina Dulantha Lalitharatne$^{2}$
\thanks{$^{1}$Department of Engineering, University of Cambridge, Cambridge CB2 1PZ, The United Kingdom
        {\tt\small \{csh66, yx388*, fi224\}@cam.ac.uk}}%
\thanks{$^{2}$School of Engineering and Materials Science in Queen Mary University of London, UK
        {\tt\small \{s.nadipineni, t.lalitharatne\}@qmul.ac.uk}}%
        \thanks{Correspondence: yx388@cam.ac.uk}
}

\begin{document}


\maketitle
\thispagestyle{empty}
\pagestyle{empty}



\begin{abstract}
Diagnostic errors remain a major cause of preventable deaths, particularly in resource-limited regions. Medical training simulators, including robopatients, play a vital role in reducing these errors by mimicking real patients for procedural training such as palpation. However, generating multimodal feedback, especially auditory pain expressions, remains challenging due to the complex relationship between palpation behavior and pain sound. The high-dimensional nature of pain sounds makes exploration challenging with conventional methods.
This study introduces a novel experimental paradigm for pain expressivity in robopatients  where they dynamically generate auditory pain expressions in response to palpation force, by co-optimizing realtime human feedback  using machine learning.  
Using Proximal Policy Optimization (PPO), a reinforcement learning (RL) technique optimized for continuous adaptation, our robot iteratively refines pain sounds based on real-time human feedback. This robot initializes randomized pain responses to palpation forces, and the RL agent learns to adjust these sounds to align with human preferences.
The results demonstrated that the system adapts to an individual's palpation forces and sound preferences and captures a broad  spectrum of pain intensities, from mild discomfort to acute distress. The study further showed that pain sound perception exhibits saturation at lower forces with gender-specific thresholds. Thus, this study presents an early attempt to use human-in-the-loop reinforcement learning to co-optimize haptic input and auditory pain expression.  These findings highlight the system’s potential to enhance abdominal palpation training by offering a controllable and immersive simulation platforms.



\end{abstract}


\section{Introduction}


While specific statistics vary by region, diagnostic errors are a universal concern. Misdiagnoses may contribute to the nearly 7 million children who die each year from preventable causes, particularly in low- and middle-income countries \cite{WHO2016DiagnosticErrors}. 
These findings underscore the critical need for systemic improvements in diagnostic processes, enhanced communication among healthcare providers, and increased patient engagement to mitigate the risks associated with diagnostic errors. Palpation is one of the primary examination methods used by physicians to examine patients in various conditions ranging from simple abdominal pain to more serious conditions such as acute appendicitis and breast, soft tissue tumors. Palpation involves multimodal feedback \cite{Robopatient_access, Robopatient_multimodal} from the patients, including facial expressions, vocal expressions such as pain sounds, gestures, and haptic feedback felt through the fingers of the physicians. Given that most of these feedback are subjective, training with diverse conditions is essential.  Robopatients, designed to represent diverse patient profiles in terms of facial expressions \cite{Robopatient_face_model, lalitharatne2021morphface, lalitharatne2022face},  and vocal expressions \cite{protpagorn2023vocal} and minimize bias, offer a potential solution to reducing misdiagnoses caused by diagnostic biases.


Pain is often the first sign of illness or injury and the pain intensity help quantify pain severity, assisting in disease assessment and treatment response \cite{manias2002observation}.
 Training robots to appropriately express pain not only deepens their capacity to respond to physical interactions but also opens new avenues for improving empathy and realism in robotic systems \cite{higgins2022robotics}.  Pain is a form of communication of distress that facilitates diagnosis and treatment in healthcare settings, creating an emotional and psychological impact. Although pain remains a mystery and consists of subjective, objective, ontological, and existential aspects \cite{dekkers2017pain}, pain recognition and generation play a significant role in many fields, including healthcare, medicine, and virtual reality. Accurate assessment of pain can significantly improve the quality of life \cite{breivik2008assessment}. Furthermore, cultural norms shape emotional experiences to varying degrees depending on factors such as the time frame, valence, and type of emotion \cite{scollon2011cultural}. These studies together  highlight the multidimensional nature of pain and its complexities beyond our current understanding.

\begin{figure*}[]
\centering
\includegraphics[width=1.5\columnwidth]{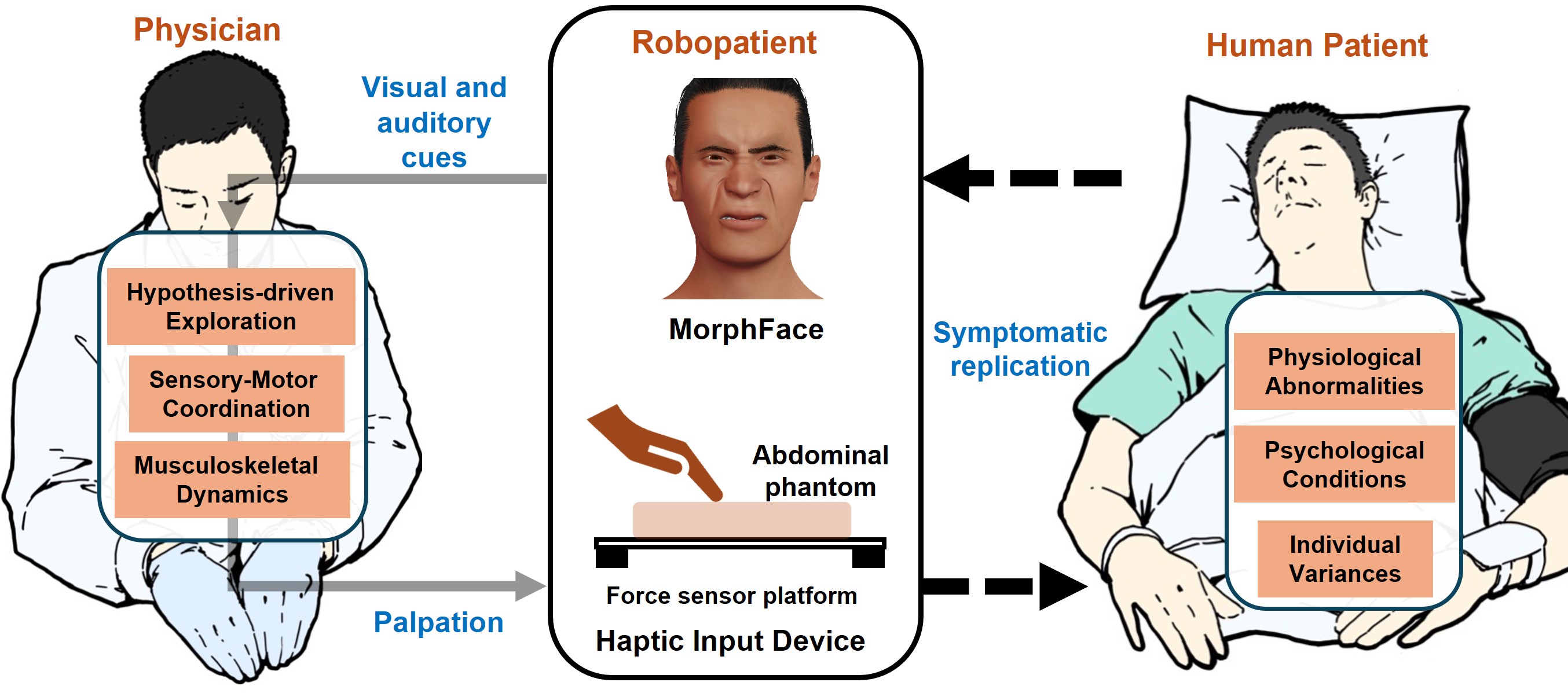}%
\caption{ Multimodal pain synthesis for robopatient. The system comprises three main components: the user (ideally a physician), the robopatient, and the human patient. The human patient under examination may exhibit various physiological abnormalities, psychological conditions, and individual characteristics, all of which influence the dimensionality of pain expression in response to palpation.
The robopatient serves as an adaptive intermediary, learning appropriate pain responses from the user feedback and aiming to replicate human pain expressions (including vocal pain sounds and facial pain expressions) realistically. During palpation, robot's abdominal phantom captures user's haptic input, which is recorded and processed through the robopatient’s internal palpation-pain mapping model. Then the robot outputs corresponding pain sounds and facial expressions through MorphFace \cite{lalitharatne2021morphface}.
The user provides feedback on the robopatient’s generated pain responses, which are iteratively refined by comparing palpation data with user feedback. This interaction loops continuously, enabling the robopatient to progressively learn and refine pain expressions while accurately capturing palpation details from the user. This concept was adapted from \cite{lalitharatne2022face} and modified to fit the new paradigm.}
\label{concept}
\vspace{-1em}
\end{figure*}


Studying the multidimensional space of human pain requires extensive human studies  and the analysis of large datasets. Conducting these studies under diverse conditions and collecting varied data is a time-intensive process, posing significant challenges in pain research.
Machine learning algorithms can significantly enhance the design of human studies by adapting in real-time to participant behavior \cite{binetti2022genetic}, which can reduce the time taken to conduct these studies. This ensures that relevant data are collected efficiently, facilitates more accurate sampling, and reduces trial times by continuously refining protocols based on intermediate results.



This study introduces a novel framework for human-in-the-loop learning by incorporating Proximal Policy optimization  to dynamically generate adaptive pain expressions in a robopatient during abdominal palpation. This work highlights early trends in how auditory expressions of pain can be shaped through interactive learning with human feedback. We used robopatient \cite{protpagorn2023vocal}, a robotic platform that dynamically generates facial expressions and sounds in response to palpation force input from human participants during abdominal palpation. In contrast to traditional studies that collect data separately for post-analysis, this experiment follows a closed-loop approach, integrating real-time human feedback into the learning process, allowing the robot to dynamically adjust its auditory pain expressions, creating a more adaptive and personalised experience based on individual palpation. Hence, this contribution can be considered an early step toward making the planning and execution of human studies on high-dimensional phenomena such as pain less time-consuming and less exhausting, by incorporating  co-optimization and real-time human feedback. The concept of robopatient in this context is illustrated in Fig. \ref{concept}. The results of the experiment demonstrated the variability of the relationship between palpation and auditory pain expression. Hence  suggesting the complexity and multidimensional nature of pain expressions, as well as the possibility that this relationship could be personalised. 

\section{Related Work}


The integration of human and ML approaches to generate adaptive auditory pain expressions in human physical twins  intersects several research areas including,
\textbf{Artificial Intelligence (AI)-Driven robopatients}, \textbf{Multimodal Human-Robot Interaction (HRI)} and \textbf{objective pain assessment using AI}.

Robopatients use AI to simulate, predict, and optimize real-world processes such as simulating patient-specific scenarios to optimize treatment plans \cite{mikolajewska2024applications}. Advancements in multimodal HRI enable communication between humans and robots using various modalities, including visual, auditory and haptic information which facilitates more natural and intuitive interactions \cite{su2023recent}, which are crucial for applications involving emotional and pain  recognition.  Recent studies on objective pain assessment using AI have explored the use of machine learning models for pain recognition by analysing physiological signals. For instance, research utilising electrodermal activity (EDA) signals has applied mixup data augmentation and support vector machines to distinguish pain levels, achieving notable accuracy improvements \cite{shantharam2024ml}. These areas collectively contribute to the  generating adaptive auditory pain expressions, enhancing the realism of robopatient.

\subsection{Pain}
Pain perception and expressions vary depending on various backgrounds such as gender \cite{rhudy2010there} and ethnicity \cite{racine2012systematic}. Infants experiencing a higher intensity of pain produce more irregular cries \cite{tiezzi2004determination} with higher amplitude \cite{lehr2007neonatal,fuller1995effect}, larger fundamental frequency variation \cite{porter1986neonatal,koutseff2018acoustic}, smaller amplitude variation, and longer duration \cite{porter1986neonatal,johnston1987acoustical}. Hence, the time scales of pain sound are also important to consider, for instance higher pitch and louder pain cries are associated with a sense of urgency \cite{porter1986neonatal, craig1988judgment}. However, when synthesizing pain artificially, design choices must ensure that the output remains within the human capacity to reliably interpret and respond—typically limited to no more than four simultaneous variables in affective or perceptual tasks \cite{BREAZEAL2003119, xxxxxx}.

Individual experience of pain can significantly differ even for the same sensory input and development of combined psychophysical or psychological models that optimize pain treatment~\cite{coghill2010individual}. Hence it makes modelling the pain sounds or vocal pain expressions challenging~\cite{lautenbacher2017phonetic, helmer2020crying}. This background shows that generating an accurate model to represent subjective pain in someone is an extremely challenging task, which requires extensive study of the human perception regarding pain expressions.

Research indicates that pitch and amplitude are primary drivers in communicating pain levels through vocalisations. Studies have further shown that during painful experiences, individuals often exhibit increased pitch and loudness in their vocal expressions. For instance, vowels such as ``u" and ``schwa," which resemble moaning or groaning, tend to have higher pitch and loudness during pain \cite{lautenbacher2017phonetic}.  Increased pitch and loudness are associated with pain vocalisations, suggesting these features are significant indicators of pain \cite{helmer2020crying}. Beyond pitch and amplitude, other acoustic features such as fundamental frequency variation, amplitude variation, and duration also play roles in conveying pain intensity. For instance, infants experiencing higher pain levels produce cries with greater fundamental frequency variation and longer durations \cite{shinya2017fundamental}. Furthermore, variations in fundamental frequency is associated with a person(an infant)'s arousal state during pain episodes \cite{goberman2013acoustics}. These studies laid the background for our selection of pain sounds and the methodology used to generate pain. 

\begin{figure*}[]
\centering
\includegraphics[width=2\columnwidth]{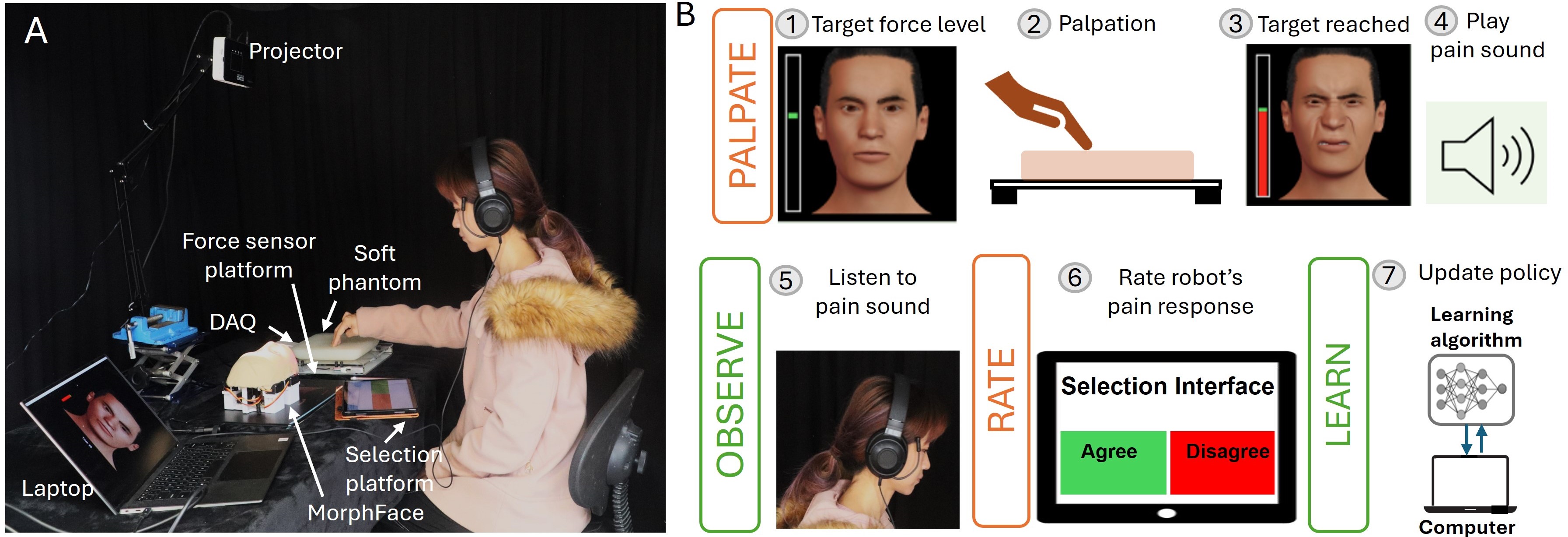}%
\caption{  Physical robot and the experiment procedure A) Robopatient setup: A scenario where the participant is palpating on  robopatient is shown. B) Data flow within the robopatient setup during the experiment. The interaction between the robot and the user starts as the user palpates on the abdominal phantom (Steps 1, 2). The force applied by the user has to reach a certain level indicated by a progress bar next to the robot's face (step 3). The progress bar is initially green and turns red when the required minimum amount of force is applied.  Based on the palpation force, the robot generates random facial expressions and sounds independent to each other (step 4). Then the user listens to the pain sounds generated by the robot for the palpation force (step 5) and provides feedback on whether they agree with the robot’s mapping of pain sounds to palpation (step 6).  This feedback updates the policy used by the PPO to generate future responses (step 7). This completes the one interaction cycle between the user and robot. Then robopatient learns the appropriate combinations of pitch and amplitude for pain sounds and optimizes its response to maximise reward. }
\label{overview}
\vspace{-1em}
\end{figure*}

\subsection{Human-ML co-optimization for emotion generation}

Learning algorithms can accommodate the multidimensionality of emotions and reduce bias, especially in the pain domain which is not well established yet in psychophysics. ML has proven effective skills in outperforming humans in analytical tasks such as bibliometry~\cite{goh2020evaluating}, emotion recognition and human studies~\cite{binetti2022genetic}.   
By employing genetic algorithms, the study in  \cite{binetti2022genetic} efficiently explored the high-dimensional space of facial expressions, demonstrating the importance of personalised approaches in understanding human emotions and improving emotion recognition systems. This study encourages moving away from stereotypical representations of emotions in research and towards approaches that capture the diversity and richness of emotional expression. In addition, prior knowledge improves our understanding of a situation, and prior experience accelerates our ability to adjust to new situations \cite{williams2013explanation}. Hence AI for adaptive experiment design has been widely discussed in design optimization \cite{busetto2013near}, managing data and knowledge \cite{zhang2022machine} in scientific discovery, identifying informative training samples \cite{tran2018combining}, benchmarking datasets and democratizing science.

Human-ML co-optimization involves combining human intuition with algorithmic optimization to enhance a robot's performance. Previous studies have used Batch Bayesian Optimization to refine robot actions through human guidance, improving task efficiency \cite{junge2020improving} and  compared human designers with model-free optimization in a hopping robot \cite{saar2018model}, highlighting how co-optimization can improve human creativity alongside ML.

Current RL algorithms allow using small, high-quality datasets instead of large, noisy data.  PPO is particularly well-suited for studying emotions and human behavior due to its efficiency in handling smaller, high-quality datasets, making it ideal for real-time, interactive environments.  Unlike other RL methods that may require vast amounts of noisy data, PPO proved to ensure stable policy and gradient updates \cite{schulman2017proximal}.  Its scalability in high-dimensional spaces, including complex domains such as emotions and human behavior, further underscores its applicability to our study, where the system must adapt to dynamic and nuanced interactions between participants and the robopatient \cite{duan2016benchmarking}.  Hence, we suggest a novel, hybrid approach for planning adaptive human experiments which typically are time-intensive and require large populations to derive useful conclusions.  Pattern recognition and analytical skills that learning algorithms possess could make a paradigm shift in how we conduct traditional human studies and how collaboration between humans and robots can accelerate learning.

\section{METHODS}

The experimental setup, shown in Fig.~\ref{overview}A, involves users palpating a soft phantom laid on a rigid platform with four force sensors, which translates the palpation force into corresponding pain expressions produced by the system (robopatient). Robopatient is equipped with this force sensor platform on which a soft phantom is placed. This soft phantom acts as the abdomen. When users palpate on this soft phantom, the force sensor platform measures the palpation force as the input and generates appropriate auditory pain expressions as outputs in real-time. We combined this output with an appropriate facial expression, which is then projected onto the morphface: robopatient's face, to make the scenario more interactive and realistic  \cite{costi2023multi}. The user listens to the pain sound generated by the robot with the headset. As the next step, the user compares his/her palpation action and the robot's pain expressions. The user then rates the congruence of these responses on a Likert scale with binary choices: \textit{Agree} and 
\textit{Disagree}. These inputs and feedback are fed to the PPO at each attempt to learn the preferences of the user. Over time, the PPO optimizes pain sounds to maximise participant feedback scores, thereby improving the alignment between input force and the robot's responses to pain. 

This process is shown in Fig. \ref{overview}B. The objective here is to maximise the perceived congruence and satisfaction of the users based on their feedback scores. Hence, we call this a human-guided adaptive framework for generating pain responses.

\subsection{Selection of Pain Sounds}

In auditory pain expressions, gender plays a significant role in shaping the acoustic properties of vocalisations. 
Social and cultural conditioning can also influence how pain sounds are expressed between genders. 
This makes expression and perception of pain sound highly dimensional and hence challenging. 

We selected three different pain sounds per gender for the experiment to observe whether users perceived certain pain sounds better than others. The pain sounds deemed appropriate to represent diverse pain intensities were selected from the pilot trials. Pain sounds were selected from the online sound effects platform (\href{https://www.soundsnap.com/}{Soundsnap}). A related study for the choice of pain sounds and facial expressions can be found in our previous study in \cite{protpagorn2023vocal}. A pair of male and female pain soundtracks used for the study is shown in Fig.~\ref{expresources}A and B. 


\subsection{Robopatient}

Robopatient consists of the following hardware discussed in detail.
The force sensor platform was made of four single-axis 20kg load cells (Haljia Inc) mounted on the four corners of a generic square shaped (250$\times$250 mm) acrylic platform, topped with a Silicone phantom (200$\times$200$\times$30 mm, Ecoflex 00-30). The distance between two adjacent load cells is 246 mm. A National Instruments DAQ (USB-6212) with MATLAB 2024a is used to acquire the signals from the load cells at a sampling rate of 1000 \textit{Hz}. The sum of the force values from each load cell is used to calculate the total palpation force $F_{tp}$. To minimise the noise in these calculations, any load cell value where the total palpation force is less than 0.5\textit{N} has been neglected based on the observations from pilot experiments. $F_{tp}$ can be calculated as follows.

\begin{equation}
  F_{tp} = \sum_{j=1}^{4} F_{j}
\end{equation}

where $F_{j}$ is the force read from the $j$th load cell mounted on each corner of the force sensor platform.



\begin{figure*}[]
\centering
\includegraphics[width=0.95\textwidth]{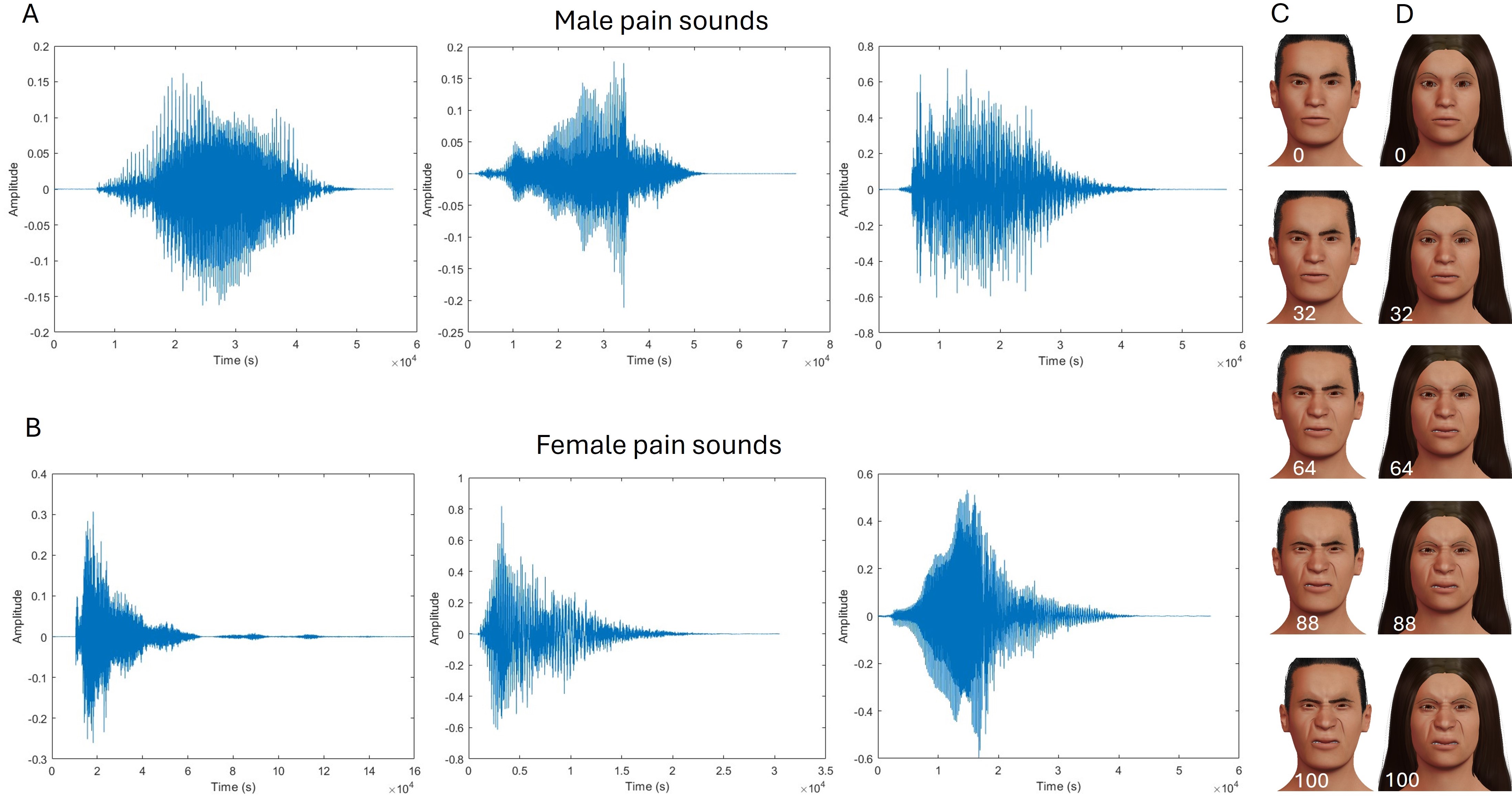}%
\vspace{-1em}
\caption{Pain expressions used in the experiment. Audio tracks to represent pain in A)  males and  B) females. These can be downloaded from \href{https://github.com/ChapaSiri12/Robopatient_PPO}{GitHub repository} where all the codes used in the study can be found. Selected frames of C)  male and D) female facial pain expressions (frame 0-neutral and frame 100-intense pain) for each gender. Selected consecutive images out of 100 were played in order to create different intensities of pain expressions for each gender are shown. }
\label{expresources}
\vspace{-1em}
\end{figure*}

MorphFace \cite{lalitharatne2021morphface} is a controllable 3D physical-virtual hybrid face that can represent facial expressions of pain displayed across diverse identities. In this case, we considered only white female and male faces to simplify the experiments and minimise the influence of other variables, such as gender and ethnic identity of the face, on pain perception, as most of our users have this background. Two sample frames showing facial pain expressions for each gender are shown in Fig. \ref{expresources}C and D.

Users selected their choices through a custom user interface (tablet) displaying two buttons: \textit{Agree} and \textit{Disagree} corresponding to `1' and `0' keystrokes respectively. An Android application (Bluetooth Keyboard \& Mouse) was used to create custom keys and connect the tablet to the computer via Bluetooth to act as an external keyboard. 


\subsection{Palpation to Pain Intensity Mapping}

For each palpation action, four parameters were changed. These are the target force level (shown in Newtons), pain soundtrack (track 1,2 and 3), amplitude  level of pain sound track (denoted as a scale), and pitch level (denoted as a scale).  These values shown below were determined empirically from a preliminary study.  

\begin{equation}
    Amplitude \ Level= [1,0.3,0.1,0.037]^T \label{amplitude_vector}
    \vspace{-1.5em}
\end{equation}

\begin{equation}
    Pitch \ Level= [0.7,0.9,1.1,1.3]^T \label{pitch_vector}
        \vspace{-1.5em}
\end{equation}

\begin{equation} \label{force_thresh}
    Force  = [5,10,15,20]^T
        \vspace{-1.5em}
\end{equation}

\begin{equation}
    Pain \ sound  track = [1,2,3]^T
\end{equation}

After recording the palpation force $  F_{tp}$, it was used to calculate the pain intensity $PI$.
\begin{equation}
    PI = \beta F_{tp}^{filtered}
\end{equation}
where $\beta=5$ is the scaling factor to map the range between 0 and 100,  $F_{tp}^{filtered}$ is the filtered output (using a moving average filter of window size 20) of $F_{tp}$.


\subsection{Dynamic Interaction optimization between the User and Robopatient}

We utilise PPO algorithm, an RL method known for its stability and efficiency in real-time environments~\cite{schulman2017proximal,henderson2018deep}. PPO is particularly well-suited for our application as it can handle continuous learning while maintaining the stability of policy updates through a clipped surrogate objective function. This mechanism prevents large deviations in policy updates, reducing the risk of destabilizing learning and ensuring consistent adaptation over time. By embedding human evaluators in the learning loop, the robot’s sound responses adapt not only to the applied force but also to the subjective preferences and interpretations of the user. Our objective in this study is to explore the feasibility of using this human–machine learning co-optimization approach to generate adaptive, expressive auditory pain responses during palpation rather than aiming to identify the best-performing algorithm.

Each time a participant palpates, the system records the magnitude of palpation force  to select an appropriate combination of facial expression and pain sound from a predefined set of options. The participant then provides the feedback score, indicating how well they agree/disagree with how well the generated pain response matches their force input. This feedback is immediately used by the system to adjust its future responses, allowing dynamic and continuous improvement throughout the interaction.

\subsection{PPO Algorithm}

Compared to alternative RL approaches, PPO demonstrates critical advantages for our human-in-the-loop paradigms: Its sample efficiency minimizes trial burden on participants given sparse binary feedback; stable convergence via clipped objectives avoids policy updates during continuous interactions. 

Soft Actor Critic(SAC)~\cite{haarnoja2018soft} offers sample efficiency but requires significant computational resources, making it impractical for real-time deployment. Deep Deterministic Policy Gradient (DDPG)~\cite{lillicrap2015continuous} and its advanced version Twin Delayed DDPG (TD3)~\cite{fujimoto2018addressing} suffer from hyperparameter sensitivity and training instability, which could disrupt human-robot interaction continuity.

Given our application involves continuous user interaction and real-time adaptation under constrained feedback, PPO provides the optimal balance of training efficiency, policy stability, and practical deployability.

The problem was formulated as an episodic Markov Decision Process (MDP), where the state at each time step is defined by the palpation force input, and the reward is based solely on the resulting response on the pain expression. Although human behavior introduces inherent stochasticity, each palpation episode is reset independently, and the user’s next input is not influenced by the robot’s action, thus satisfying the  assumptions of an MDP.

As identified earlier, using a  clipped objective function in the PPO prevents large updates that could destabilize the training process, ensuring that policy updates remain within a reasonable range and promoting stable learning. Additionally, PPO maintains and updates both the policy network and the value function network. The policy network determines the actions to take based on the current state, while the value function network estimates the expected return, aiding in the improvement of the policy's decision-making process.

The PPO algorithm maintains two neural networks: a policy network, which selects actions (selecting appropriate pitch and amplitude) based on the palpation force, and a value network, which estimates future rewards. The policy is iteratively updated through mini-batch gradient descent, optimizing the clipped objective to balance learning stability and performance. Concretely, we use separate actor/critic MLPs with two hidden layers of $64$ units and \textit{tanh} activations, optimized with Adam (learning rate $3\times110^{-4}$), minibatch size $64$, and $10$ epochs per update; these choices follow the common PPO practice~\cite{schulman2017proximal,henderson2018deep}. The pilot runs showed the fastest stable convergence, whereas training was slower at $1{\times}10^{-4}$ and oscillatory at $1{\times}10^{-3}$. 

\begin{figure}[!b]
\centering
\includegraphics[width=0.9\columnwidth]{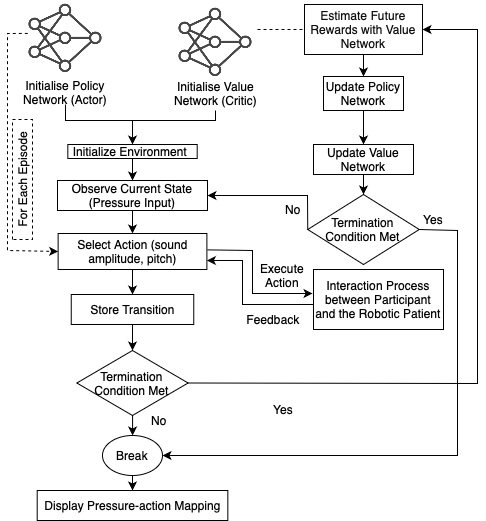}%
\caption{Process flow diagram of the PPO algorithm.} 
\label{flowchart}
\vspace{-1em}
\end{figure}

The workflow of PPO used in this study is illustrated in Fig.\ref{flowchart}. The algorithm starts by initialising its policy and value network as it sets up the interaction environment involving the robot and the user. For each interaction, the algorithm observes the current palpation data and selects a sound combination (pitch and amplitude) using the policy network. The selected action is then presented to the user, who provides binary feedback (\textit{Agree} or \textit{Disagree}), which is taken as a reward. The system updates its state based on the feedback and stores transitions (state, action, reward, next state) for policy and value updates. PPO optimizes the policy network, while the value network is updated to improve future rewards. Then it checks for a termination condition after each episode, and if met, the episode ends, displaying the most successful action-input mapping. If the condition is not met, the system continues learning; otherwise, the environment resets and starts a new episode. We terminate when the last 10 iterations contain $\geq 8/10$ \textit{Agree} responses. This $80\%$ threshold was selected in pilot sessions as the most stable stopping rule that avoids premature convergence while limiting participant fatigue.


A detailed  workflow of applying PPO is summarized in the following steps:
\begin{enumerate}
    \item \textit{Initialization:} The system initializes its policy and value networks, which are neural networks responsible for deciding the appropriate facial expression and combinatorial sound parameters to respond with, based on the palpation force (input) from the participant. The environment, which consists of the robopatient and the participant, is also initialized.
    \item \textit{Action Selection:} For each interaction, the system observes the current input and uses the policy network to select an action, which is a combination of a facial expression and sounds parameters.
    \item \textit{Execution and Feedback:} The selected action is presented to the participant, who provides feedback on how appropriate they found the response. This feedback is a score from 1 to 4.
    \item \textit{Reward and Observation Update:} The system receives the feedback score as a reward and updates the current state based on the new force input. The transition (current state, action, reward, next state) is stored for policy and value updates.
    \item \textit{Policy and Value Network Update:} Using the collected transitions, the PPO algorithm computes the advantage estimates and updates the policy network by optimizing the clipped surrogate objective function. Simultaneously, the value network is updated to better predict the expected future rewards.
    \item \textit{Termination Check:} The system checks if the termination condition is met (currently set to the probability of achieving a feedback score of `strongly agree' for ten consecutive interactions to $80\%$). If the condition is met, the episode ends, and the system displays the mapping between actions and force inputs that led to the highest participant satisfaction. 
    \item \textit{Reset and Continue:} If the termination condition is not met, the system continues to the next iteration. If it is met, the environment resets, and a new episode begins.
\end{enumerate}



\section{Experiment}

Five healthy participants (2 males, 3 females) aged between 25 and 46 years (mean-34.5, SD-11.77) from the University of Cambridge, with no prior experience in pain to palpation mapping and no history of sensorimotor impairments were chosen to participate in this experiment. Participants were chosen so they represent diverse backgrounds such as ethnic (European-2, Asian-2, and American-1), gender (male, female), and technical and non-technical experience to capture a wide range of perceptual responses and interaction styles. This diversity introduces natural variability into the system, helping to evaluate the generalizability of the proposed approach and simulate realistic user scenarios during the experiment.

According to the experiment setup shown in Fig. \ref{overview}A, when a trial (palpating, observing robot's expressions and providing user feedback) begins, the participant will be presented with a graphical interface. This interface comprises of   a face from a chosen gender projected on the MorphFace. This face changes its expressions from mild to intense based on pain intensity. The green colour progress bar  appears when force is applied on the force sensor platform and  turns red as the required force level is reached. The length of the bar is mapped approximately to the force applied. The maximum target force from \ref{force_thresh} levels was 20\textit{N}, hence this was the maximum length of the bar. The four target force levels in \ref{force_thresh} were introduced in the study to get diverse palpation inputs from participants. 

During the experiment, each participant was asked to palpate the abdominal phantom until they reached the target force level. A period of 5 s is given for the palpation. A pain sound was played via the headphones (Razer Kraken X Lite) when they reached or exceeded the target force level and its amplitude and pitch were determined by the PPO agent. Then the projection displays ``Select" indicating the user to make a decision by clicking the options on the tablet.   The participants are given 3 s to decide whether they ``agree'' or ``disagree'' with the robopatient's pain response. After this, the next trial begins. Then the participant selected their choice: whether they \textit{Agree (0)} or \textit{Disagree (1)} on the auditory pain expression played given their palpation through the tablet. However, when participants were unable to make a decision within the given time frame, we recorded their choice as 0.5 to avoid bias.

 A random pain sound and force threshold combination will be selected for the PPO training. The PPO agent selects a random amplitude and pitch value from \ref{amplitude_vector} and \ref{pitch_vector} for the first trial. The PPO agent changes these values based on user input. 120 trials were conducted per participant for the pain expressions of each gender. Therefore, there were a total of 240 trials per participant during the experiment.


Participants were allowed to familiarise themselves with the system with six familiarisation trials per pain expressions from each gender. This data was not used for learning.
Counterbalancing is introduced to this study by changing the order of the expressions the participants began with. For example, if a participant started the male experiment first, the next participant started the female experiment instead. This is to reduce any bias in the experiments.

\begin{table*}[]
\caption{4-point Likert scale questionnaire and answers provided by the participants (0 = strongly disagree and 3 = strongly agree) }
\vspace{-1em}
\centering
 \small
\resizebox{0.8\textwidth}{!}{%
\begin{tabular}{lc}
\hline
\multicolumn{1}{c}{\textbf{Questions}} & \textbf{Median answer} \\ \hline
1. Were the vocal pain expressions realistic? & 2 \\
2. Were the vocal pain expressions clearly distinguishable? & 2.5 \\
3. Was the variation of pain sounds adequate? & 2 \\
4. Was the time taken for the experiment reasonable? & 2 \\
5. Which attribute you paid attention to first:  facial expression or pain sound or both? & Pain sound \\ \hline
\end{tabular}%
}
\label{questionnaire}
\end{table*}



\begin{figure}[!b]
\centering
\includegraphics[width=0.495\textwidth]{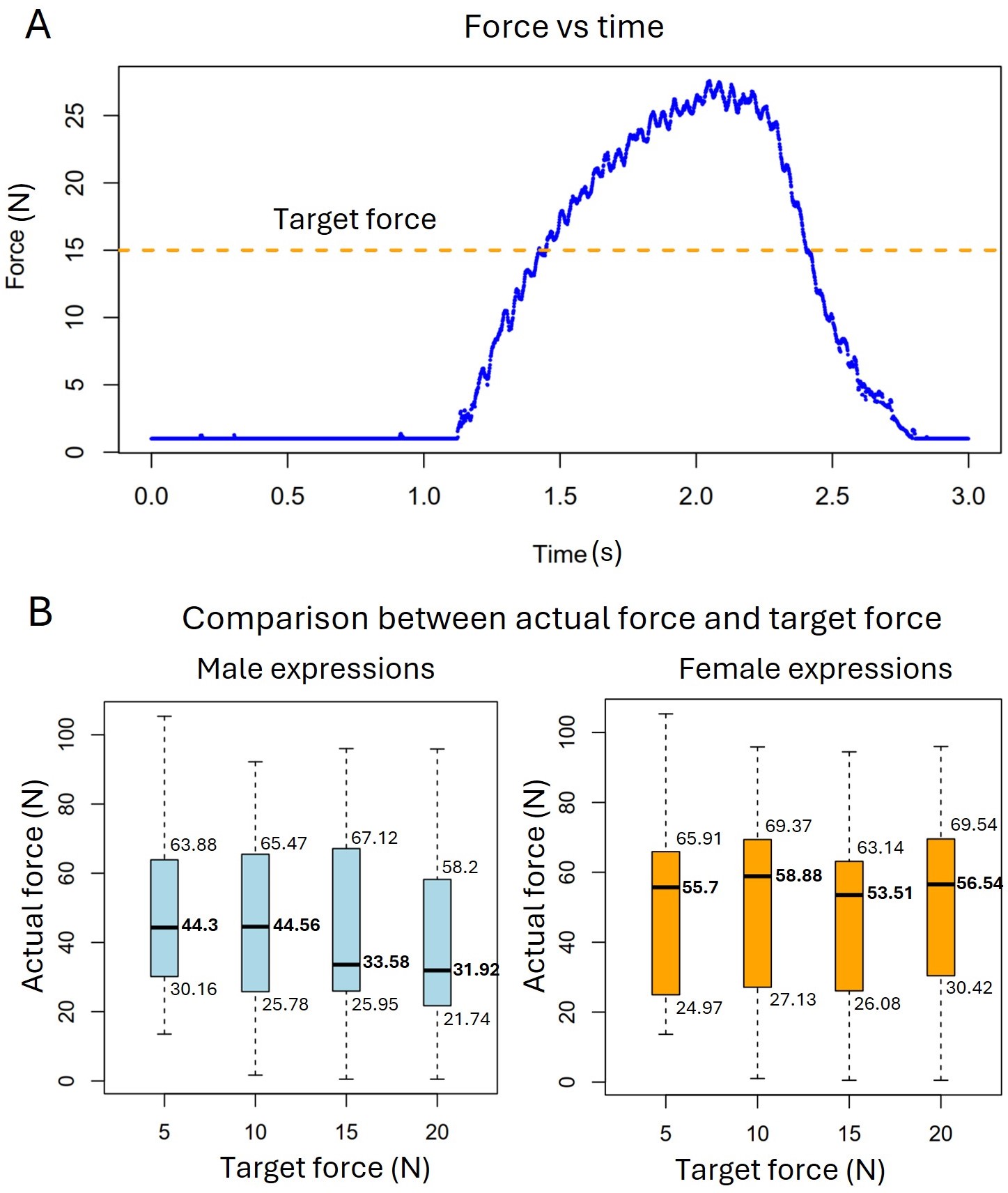}%
\vspace{-.5em}
\caption{ A) The force distribution for a given attempt for palpation is plotted against time. This corresponds to the target force of 15N. B) Box and whisker plots indicate the spread of actual palpation forces observed for each target force for the robot's pain expressions of each gender. }
\label{Forceplot}
\end{figure}

\begin{figure*}[]
\centering
\includegraphics[width=.95\textwidth]{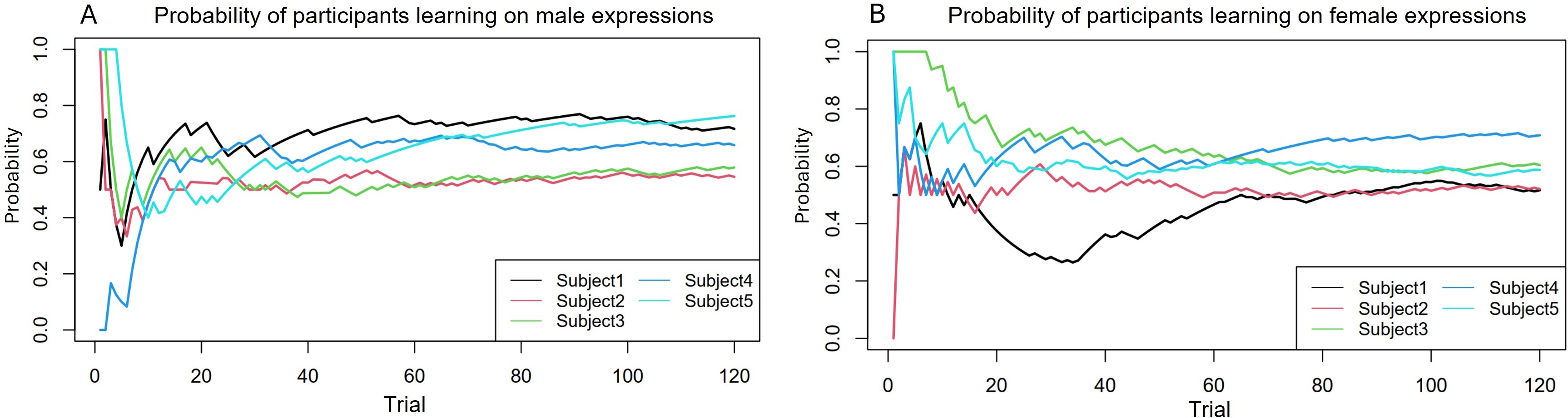}%
\vspace{-1em}
\caption{Cumulative probability at each stage of learning trial is plotted against each type of sound: A Male sound B Female sound. }
\label{Learning_all}
\end{figure*}

\subsection{Post Study Questionnaire}

We presented five Likert scale questions followed up by an open ended question to obtain additional thoughts on the experiment. The objective was to assess user satisfaction, identify usability issues, and collect suggestions for future improvements. Participants were given the options from 0-3 (worst-0, best-3) to answer these questions. The questionnaire consists of questions covering variable demographics and usability. These questions can be found in TABLE \ref{questionnaire}.

\section{Results and Discussions}

Fig. \ref{Forceplot}A indicates the force distribution during palpation ( period within 5 s) in a single trial. The target force level is marked.  Fig. \ref{Forceplot}B illustrates the variance between the target and recorded maximum force in each trial for both male (p=4.54e$^{-136}$ $<0.05$) and female (p=5.88e$^{-164}$ $<0.05$) pain expressions. While there were variations in the range of forces we could observe for each target force level, average actual forces converged around 40 and 60 N for male and female pain expressions, respectively. These trends could have  often been influenced by participants' subjective comfort.


Fig.~\ref{Learning_all} shows the cumulative probability trends considering 120 trials per gender per participant, which indicates the adaptability of participants in learning pain expressions across different conditions. The early trials demonstrate variability, a hallmark of the system’s exploration phase, where different combinations of pain sounds and expressions are tested. As the trials progress, PPO’s optimization process leads to increased stability and consistency, with participants’ preferences becoming more predictable and the system converging toward higher probabilities of selecting optimal responses.  PPO’s ability to adapt in real-time is reflected in the relatively steady learning curves. These learning curves show how participants’ feedback is incorporated into the system to improve the accuracy of selected pain expressions. In addition, relatively smoother trajectories in some participants from the beginning compared to others may be explained by individual differences in feedback interpretation. One key insight is that the variability in pain expression among participants highlights the importance of having a flexible, adaptive learning model. Some individuals may show little reaction to discomfort due to high pain tolerance, while others may react strongly even to minor stimuli. PPO’s ability to adjust to these individual differences is reflected in the distinct learning curves of participants. 
These show that the system can handle varying feedback with sample efficiency. PPO’s capacity to adjust policies dynamically allows the system to account for these variances, resulting in distinct learning trajectories for different individuals. 

\begin{figure*}[!t]
\centering
\includegraphics[width=\textwidth]{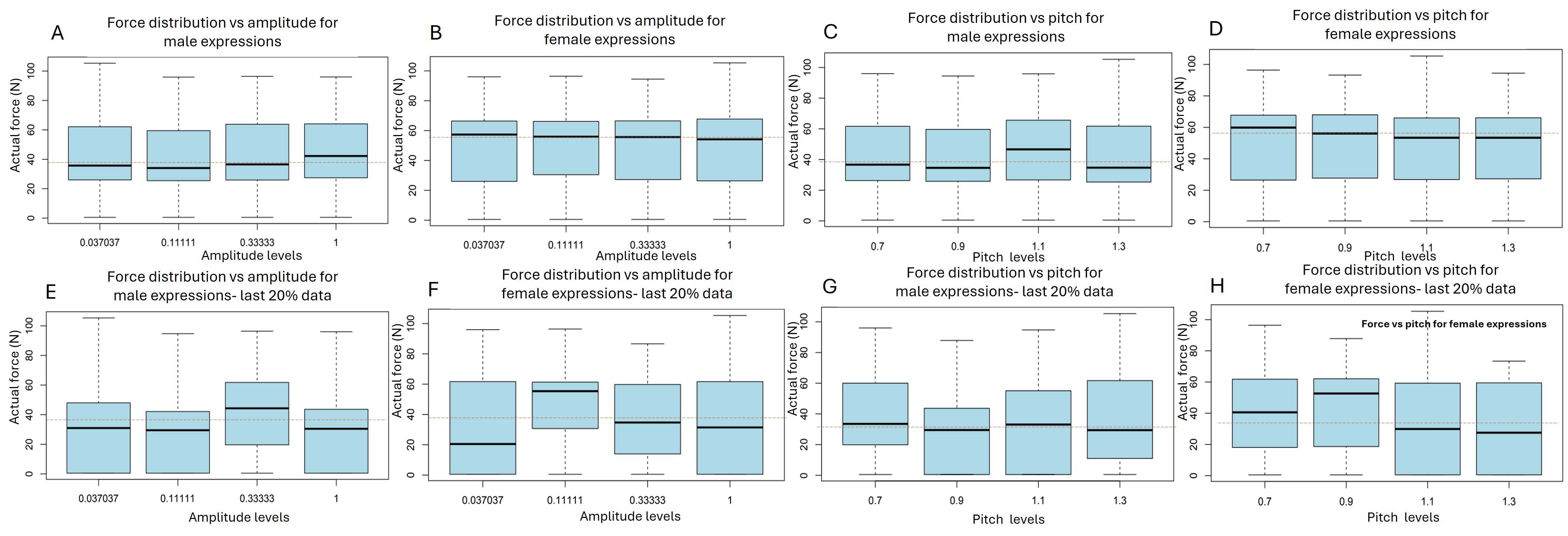}%
\vspace{-0.5em}
\caption{A-B plots the actual force  distribution for different amplitude levels for the pain expressions of each gender. Similarly, C-D plots the actual force  distribution for different pitch levels for the pain expressions of each gender. Similarly, last 20\% actual force has been plotted against amplitude and pitch levels for pain expressions of a given gender in E-H. }
\label{Soundplots}
\end{figure*}


\begin{figure*}[]
\centering
\includegraphics[width=\textwidth]{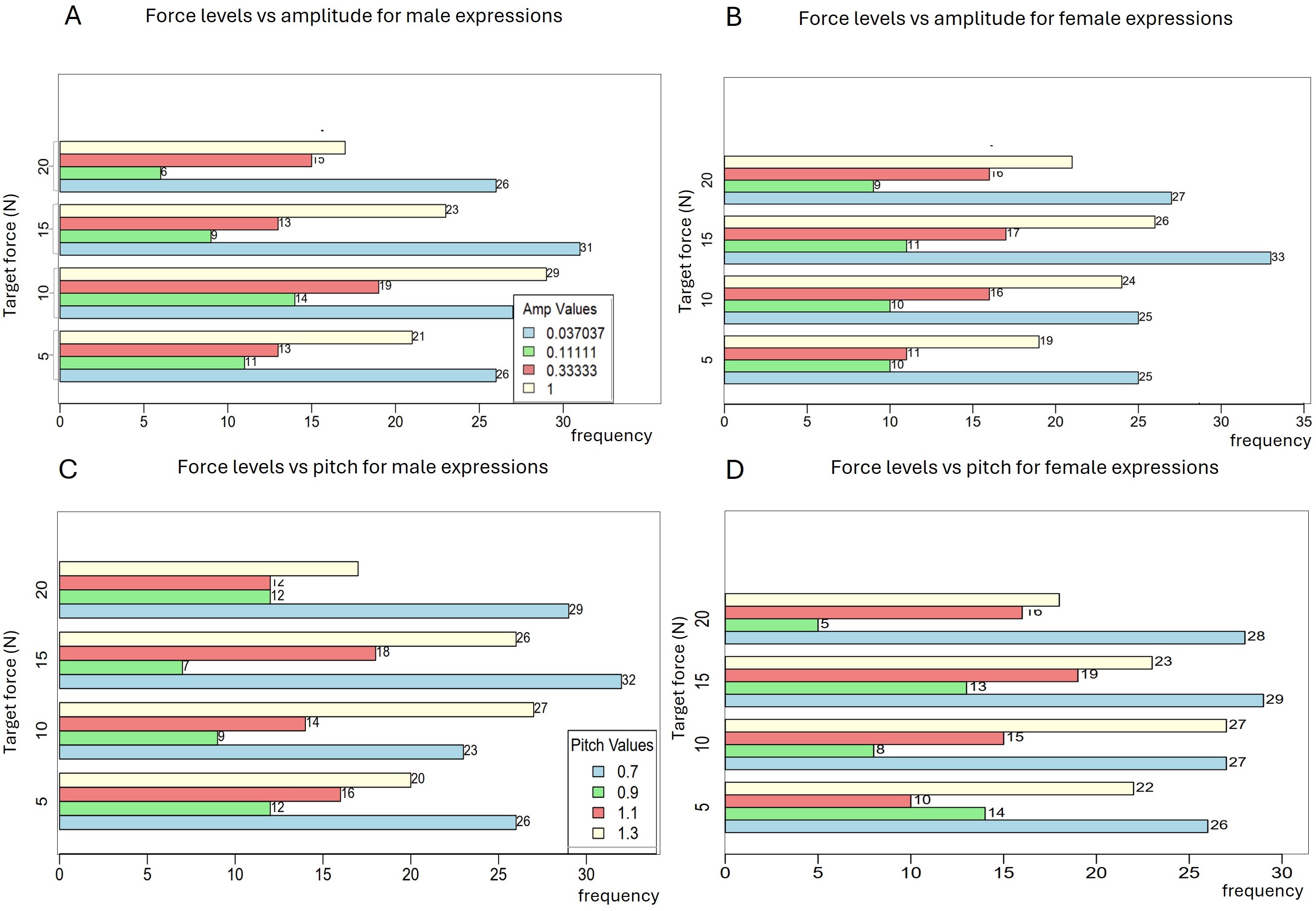}%
\vspace{-0.5em}
\caption{ The frequency of choices (modes) for the amplitude and pitch for a given force level is plotted from A-D. A and B illustrate participant choices for each force level vs amplitude of the pain sound for male and female pain expressions respectively. C and D illustrate participant choices for each force level vs pitch of the pain sound for male and female pain expressions respectively.  }
\label{Overallplots}
\vspace{-0.75em}
\end{figure*}

We analyzed pain expression responses separately for male and female vocalizations to examine participant trends across genders. As shown in Fig.\ref{Soundplots}A and B, the actual palpation forces for all target force levels averaged around 40 N when participants responded to female pain sounds (p=1.421e$^{-5}$ $<$0.05). A similar pattern was observed for male pain sounds (Fig.\ref{Soundplots}C and D), but with a higher average palpation force of approximately 60 N, suggesting that participants tolerated or applied greater pressure when hearing male pain expressions. Notably, this trend persisted across both pitch and amplitude variations.

However, when focusing on the final 20\% of palpation attempts (Fig.~\ref{Soundplots}E–H), the distinction between male and female pain sound responses diminished. In this phase, participants converged on an average force of around 40 N across all conditions (p=0.374$>$0.05), regardless of gender or acoustic parameters. We interpret this convergence as part of a learning or calibration process, where participants adjusted their behavior to a self-determined comfort zone over time.

The observed gender-based variations in initial responses may be explained by both physiological and psychosocial factors. Prior research suggests that emotional salience and perceived vulnerability differ based on voice gender; for example, female vocalizations may be interpreted as more urgent or emotionally expressive \cite{schirmer2006beyond}, leading to more cautious palpation. In contrast, male voices may be perceived as more tolerant or stoic, resulting in participants exerting higher forces. These findings align with literature in social cognition and gendered perception of pain, which suggests that listener biases and social norms can influence perceived severity and response behavior \cite{soetanto2006there, wiesenfeld2005sex}.

These results underscore the need for personalized and adaptive pain models in robopatient systems, especially when integrating human-in-the-loop feedback. Future work will explore individual and cultural biases more deeply and examine how robot voice characteristics modulate user behavior in procedural tasks.


Although these trends are  visible, it is possible that participants’ preferences may reflect personal comfort more than an accurate match between force and pain sounds. This introduces a potential bias where participants might not always map their objective experience of pain to the robopatient’s output. Given these uncertainties, adaptive learning focusing on individuals could be a better alternative to generating pain expressions based on large datasets, where robots directly mimic human pain sounds rather than relying on general subjective interpretations. 

We then calculated the frequency of choosing each pitch and amplitude level for a given palpation force for each trial. This is illustrated in the bar charts from Fig. \ref{Overallplots}A-D. Each target force level has 4 bars representing the frequency of each parameter (pitch or amplitude), corresponding to each pitch/amplitude level.


\begin{figure*}[]
\centering
\includegraphics[width=0.7\textwidth]{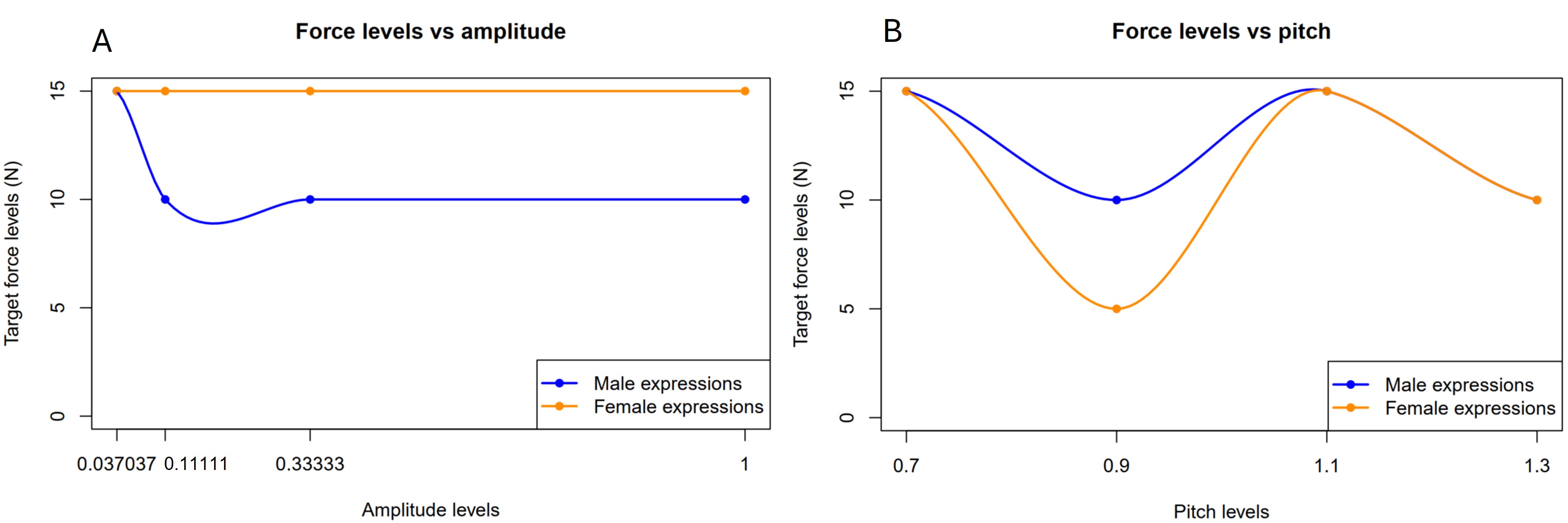}%
\vspace{-0.5em}
\caption{The modes corresponding to each amplitude or pitch values shown in Fig. \ref{Overallplots} are plotted against A) amplitude and B) pitch for male and female pain expressions respectively. }
\label{results}
\vspace{-0.75em}
\end{figure*}

The frequency of choices for the pitch, when each gender is considered, seems to follow the same trend as shown in Fig. \ref{Overallplots}C-D. The modes for each amplitude level have been plotted separately against the corresponding amplitude and pitch in Fig. \ref{Overallplots} A-D in Fig. \ref{results}A and B respectively. It can be seen that lower amplitudes were preferred for higher target forces when the force was between 5-15 N, and a higher amplitude was preferred for higher forces when the target was set to 20 N. The force preferences when the amplitude of sound increases both male and female expressions, in this case, were 15, 10, 10, 10 and 15, 15, 10, 15 respectively.

Curves had opposite trends in amplitude for each gender while the trend was similar for pitch for both genders. 
However, although we would have assumed that the amplitude and pitch of the pain sound might be proportional to the palpation force, we observed a sharp decline of palpation force (5N) around 0.9 pitch level for both genders. Hence, this can be considered proof of the multidimensional nature of pain sounds in palpation. 
The trend  for these two occasions when pitch is considered are somewhat similar. Hence, we observe a pattern in the pitch and amplitude of pain sound corresponding to palpation forces. However, we argue that finding a general equation to generate appropriate sound properties in pain requires more trials covering a larger population. Again, these trends were often influenced by participants' subjective comfort. This suggests that reverse correlation, where participant feedback may not always align with the objective experience of pain, can complicate the accurate generation of pain expressions. Additionally, cognitive biases or emotional factors could influence how pain sounds are processed, leading to less consistent responses. These findings underscore the importance of accounting for gender-based perceptual differences in studies involving pain sounds, as
they can significantly affect the interpretation of results and the design of such experiments.

Results of the experiment conclude that there is a complex and multidimensional relationship between palpation force and auditory pain expressions. While general trends were observed—such as lower amplitudes being preferred for lower forces and higher amplitudes for higher forces—participants’ subjective comfort often influenced their feedback. This introduces potential bias in the mapping of pain sounds, indicating that reverse correlation may complicate the alignment between the objective experience of pain and the robot's expressions. The study highlights the challenges in accurately generating pain sounds. It suggests that adaptive, real-time learning models may be more effective than relying on predefined datasets for pain expression generation. Hence, co-optimizing with machine learning may offer a more adaptable approach to studying multidimensional human data.

Our experimental results revealed gender-related differences in the perception and expression of pain sounds, particularly in the thresholds at which pain was perceived as intense. While this study did not aim to investigate these differences in depth, prior literature suggests that both physiological and psychosocial factors may contribute. 

On the physiological side, differences in pain sensitivity, hormonal modulation (e.g., estrogen’s influence on nociceptive processing), and auditory perception thresholds have been reported across genders \cite{fillingim2009sex}. Psychologically, gender norms and expressivity may influence how participants perceive and evaluate pain sounds—e.g., males may downregulate perceived intensity, while females may be more attuned to emotional vocalizations \cite{wiesenfeld2005sex, soetanto2006there}. Such variations may persist across gender, age, and ethnicity. However, broader observations involving diverse participants and robotic systems will be necessary to generalize these findings within the domain of human–robot interaction.

\subsection{Post Study Questionnaire}

For all the questions user feedback averaged around `agree', except for question 2 where the median answer was between `agree' and `strongly agree'. 
Qualitative post-experiment feedback revealed that participants focused more on vocal pain expressions and this could be due to its discrete nature.  The range of maximum pain intensity associated with vocal expressions may also contribute to this outcome, as participants might rely more on vocal cues to interpret pain levels. 

Participants suggested enhancing the robopatient simulator by adding diverse vocal pain expressions, such as long grunts and screams, and by developing more comprehensive palpation training programs. Some further preferred more time for decision making. Under additional comments from the participants, were suggestions to incorporate the speed of palpation to map appropriate pain sounds, that it was easier to distinguish female pain sounds compared to that of a male and it would have been useful to know more about a certain injury and the type of pain to have more resemblance to the scenario. 

\subsection{Limitations, Learning Perspective and Challenges }
While the study yielded positive findings, several limitations exist. The pain-to-sound mapping was limited to three levels, making it difficult for participants to differentiate feedback at maximum pain intensity. Additionally, the small sample size of naive participants may have reduced the statistical power, resulting in insignificant outcomes. This study is presented as a pilot to demonstrate how ML can make emotion-based human experiments less time consuming and more tailored for individual differences. This approach supports robot learning as well as the integration of vocal pain feedback into robopatients for abdominal palpation training. We plan to expand our investigations with a larger number of participants from diverse genders and ethnic backgrounds and participants with different palpation experience levels to address
these limitations.  In addition, we intend to expand the experiments to medical professionals and patients with different types of pain as well.

The design of adaptive experiments presents unique challenges, particularly in terms of prediction accuracy and the expansive search space of possible actions and outcomes. In environments where feedback from human participants drives decision-making, an accurate and well-calibrated prediction model is essential. This is to ensure that the system can effectively respond to new inputs. Without precise predictions, the estimated utility of actions or information gained from feedback becomes unreliable, leading to suboptimal learning outcomes.

RL, particularly PPO, addresses these challenges by providing a robust framework for continuous adaptation in dynamic, high-dimensional spaces such as human emotions and pain expressions. PPO’s advantage lies in its ability to balance exploration and exploitation effectively, ensuring that the system learns from its actions while continuing to test new strategies. This balance is critical when the search space for potential experiments, such as different pain expressions, is vast. The PPO algorithm iteratively refines its policy through clipped objective functions, which help prevent drastic deviations from previously learned strategies, ensuring stable learning over time.

Two main challenges in adaptive experimental design are, obtaining the accuracy of the prediction model and the large search space of possible experiments. If the prediction model is not accurate, then the estimates of information gained, and experimental utility would be affected. Hence accurate and well trained and calibrated prediction models are important. In many applications, the set of possible experimental designs is very large- hence need better ways to efficiently search the design space using ML. High dimensionality in concepts such as pain could make this even more tedious.

For the experimental design, it is important to consider that some people may have a high pain tolerance and may not exhibit obvious signs of pain, while others may have a low pain threshold and show signs of pain even with minor discomfort. Additionally, ethnic, cultural and individual differences can impact how people express and perceive pain. When reading about pain in someone, it is essential to take a holistic approach and consider all factors including timescales. Additionally, various techniques for generating pain beyond palpation, such as pinching, will be considered.

While this work  considers palpation using discrete force levels, real-world palpation occurs in a continuous force space. To better reflect this, we selected PPO, which is well-suited for continuous action spaces. This choice is made to facilitate future extensions toward modeling palpation as a continuous control task. Furthermore the current study does not aim to benchmark reinforcement learning algorithms, it introduces a novel paradigm for human–ML co-optimization in affective robotics. Our focus is on demonstrating feasibility and exploring trends rather than establishing algorithmic superiority. Future work will include comparative evaluations with alternative RL and data-driven methods to assess model generalizability and performance.

\section{CONCLUSIONS}

Our study highlights how machine learning can be used by a robot to dynamically adjust auditory pain expressions based on palpation force. This optimizes responses based on palpation effort and feedback, which can be adopted to co-optimize human responses in human studies in general.  The results of the experiment conclude that there is a complex, multidimensional relationship between palpation force and auditory pain expressions. General trends were observed—such as the existence of gender differences in pain vocalisations, where a lower pitch and consistent amplitude generally characterised male pain sounds. In comparison, female expressions showed higher
pitch and more variable intensity. This introduces potential bias in the mapping of pain sounds, indicating that reverse correlation may complicate the alignment between the objective experience of pain and the robot's expressions. The study highlights the challenges in accurately generating pain sounds and suggests that adaptive, real-time learning models may be more effective than relying on predefined datasets for pain expression generation. Despite the limitations, this study lays the groundwork for using reinforcement learning to personalise pain-palpation mapping and improve the accuracy of robopatient simulations in medical training. Moreover, our approach could be a cornerstone study for mapping the relationships between other feedback modalities such as visuo-haptic, visuo-auditory and many other combinations. Further studies should aim to incorporate diverse populations to generalise the findings and refine the accuracy of palpation-pain sound mapping that can create a significant impact on high-fidelity physical twins. 
While PPO provide ideal for out interaction constraints, future work could benchmark newer RL methods in scaled deployments with richer feedback modalities.

\section*{ACKNOWLEDGMENT}

We thank all participants for their invaluable contributions to this study. Each participant received a £10 Amazon voucher as a token of appreciation.

\section*{Supplementary Materials}
The supplementary materials for this study are available on our \href{https://github.com/ChapaSiri12/Robopatient_PPO}{GitHub repository}.

\section*{Ethics Statement}
The studies involving humans were approved by the Ethics Committee of the Department of Engineering, University of Cambridge, United Kingdom under the light touch / low-risk scheme (No.159, 14/04/2023). The studies were conducted by the local legislation and institutional requirements. The participants provided their written informed consent to participate in this study.

\balance
\bibliography{IEEEexample}
\bibliographystyle{IEEEtran}
\end{document}